\documentclass[letterpaper, 10 pt, conference]{ieeeconf} 

\usepackage[utf8]{inputenc}
\usepackage[T1]{fontenc}
\usepackage{acro}
\usepackage{amsfonts}
\usepackage{amsmath}
\usepackage{graphicx}
\usepackage{pgfplots}
\usepackage{pgfplotstable}
\usepackage[separate-uncertainty=true]{siunitx}
\usepackage{subcaption}
\usepackage{tikz}
\usepackage{url}

\IEEEoverridecommandlockouts
\usetikzlibrary{arrows, calc, 3d, scopes}
\pgfplotsset{compat=1.10}
\newcommand*\circled[1]{\tikz[baseline=(char.base)]{\node[shape=circle, draw, inner sep=2pt, minimum size=14pt] (char) {#1};}}

\DeclareAcronym{BN}{short=BN, long=batch normalization}
\DeclareAcronym{MDP}{short=MDP, long=Markov decision process}
\DeclareAcronym{NN}{short=NN, long=neural network}
\DeclareAcronym{RL}{short=RL, long=reinforcement learning}
\DeclareAcronym{PPH}{short=PPH, long=picks per hour}
\DeclareAcronym{UCB}{short=UCB, long=upper confidence bound}
\DeclareAcronym{TCP}{short=TCP, long=tool center point}

\DeclareMathOperator*{\argmax}{arg\,max}

\title{\LARGE \bf Robot Learning of Shifting Objects \\ for Grasping in Cluttered Environments}

\author{Lars Berscheid, Pascal Meißner, and Torsten Kröger%
\thanks{Intelligent Process Automation and Robotics Lab (IPR),
	Karlsruhe Institute of Technology (KIT)
	{\tt\small \{lars.berscheid, pascal.meissner, torsten\}@kit.edu}}%
}

\begin{document}

\maketitle

\thispagestyle{empty}
\pagestyle{empty}

© 2019 IEEE. Personal use of this material is permitted. Permission from IEEE must be obtained for all other uses, in any current or future media, including reprinting/republishing this material for advertising or promotional purposes, creating new collective works, for resale or redistribution to servers or lists, or reuse of any copyrighted component of this work in other works.
\vspace{5mm}

\begin{abstract}
Robotic grasping in cluttered environments is often infeasible due to obstacles preventing possible grasps. Then, pre-grasping manipulation like shifting or pushing an object becomes necessary. We developed an algorithm that can learn, in addition to grasping, to shift objects in such a way that their grasp probability increases. Our research contribution is threefold: First, we present an algorithm for learning the optimal pose of manipulation primitives like clamping or shifting. Second, we learn non-prehensible actions that explicitly increase the grasping probability. Making one skill (shifting) directly dependent on another (grasping) removes the need of sparse rewards, leading to more data-efficient learning. Third, we apply a real-world solution to the industrial task of bin picking, resulting in the ability to empty bins completely. The system is trained in a self-supervised manner with around \num{25000} grasp and \num{2500} shift actions. Our robot is able to grasp and file objects with \num{274 \pm 3} picks per hour. Furthermore, we demonstrate the system's ability to generalize to novel objects.
\end{abstract}

\section{INTRODUCTION}

Grasping is an essential task in robotics, as it is the key to successfully interact with the robot's environment and enables further object manipulation. The fundamental challenges of grasping are particularly visible in bin picking, the task of grasping objects out of unsystematic environments like a randomly filled bin. It emphasizes challenges as partially hidden objects and an obstacle-rich environment. Furthermore, bin picking is of enormous significance in today's industrial and logistic automation, enabling pick and place applications or automatic assembly. To enable future robotic trends like service or domestic robotics, bin picking and therewith robotic grasping needs to be solved robustly.

In general, grasping is more complex than a single clamping action. For example, surrounding obstacles might prevent all possible grasps of a specific object. In this case, it needs to be moved first so that it can be grasped afterwards. While pre-grasping manipulations are trivial for humans, they require interactions like sliding, pushing or rolling, which are complex for robots. In the context of bin picking, pre-grasping is essential to empty a bin completely, since the bin itself might block grasps in its corners. Additionally, when items are stored as space-efficiently as possible, objects often prevent each other from being grasped in densely filled bins.

\begin{figure}[t]
	\center
\begin{tikzpicture}
	\node[anchor=south west, inner sep=0] (overall) at (0,0) {\includegraphics[trim=115 40 0 40, clip, width=0.85\linewidth, angle=-90, origin=c]{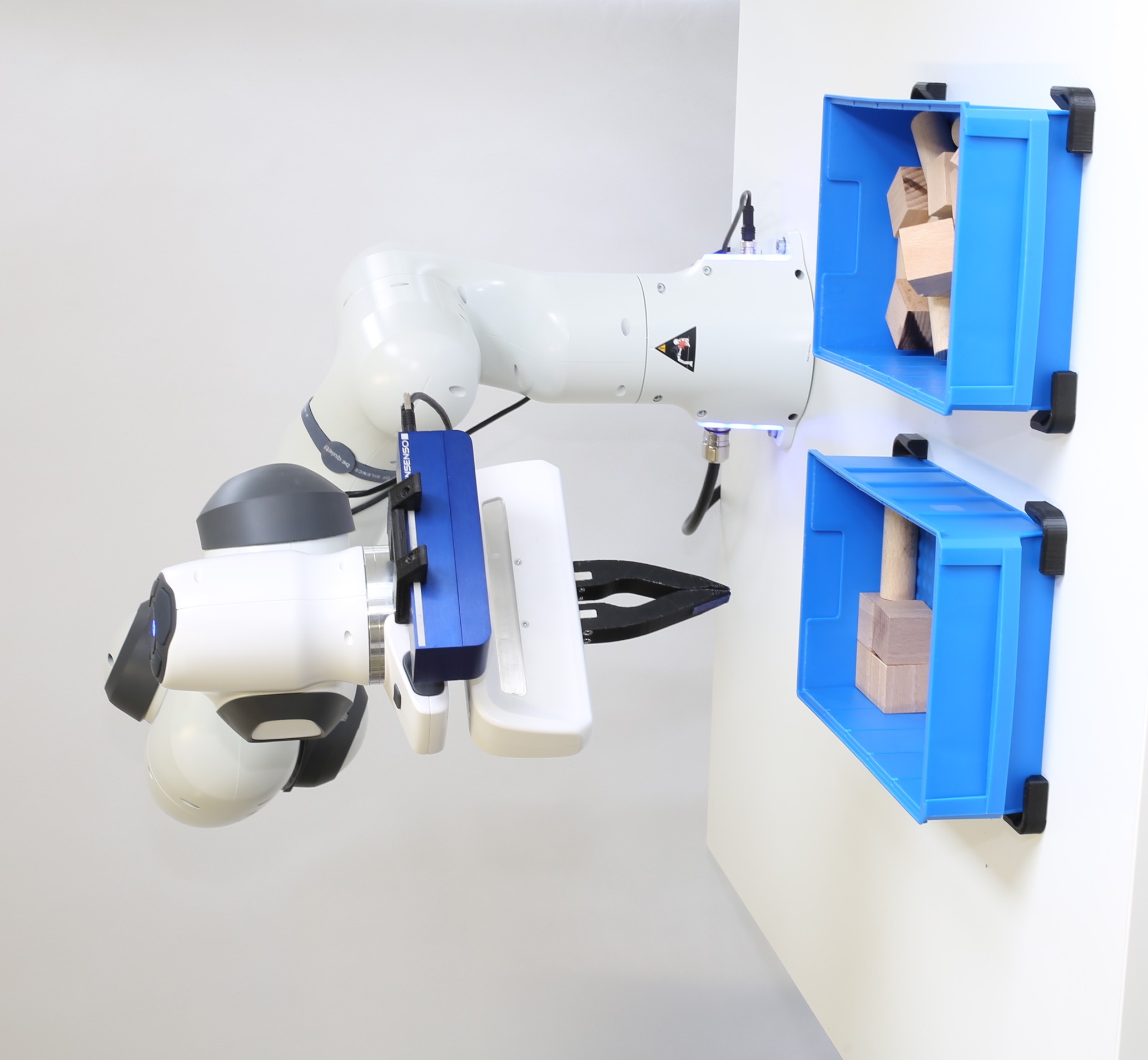}};
	\begin{scope}[x={(overall.south east)}, y={(overall.north west)}]
		\node[] at (0.88, 0.5) {\circled{a}};
		\draw [->] (0.846, 0.48) to (0.68, 0.36);
		
		\node[] at (0.12, 0.75) {\circled{b}};
		\draw [->] (0.152, 0.734) to (0.42, 0.65);
		
		\node[] at (0.12, 0.5) {\circled{c}};
		\draw [->] (0.152, 0.49) to (0.44, 0.4);
		
		\node[] at (0.1, 0.3) {\circled{d}};
		\draw [->] (0.135, 0.29) to (0.38, 0.23);
	\end{scope}
	
	\node[] (1) at (0.2, -0.7) {\includegraphics[trim=300 100 300 100, clip, width=0.29\linewidth]{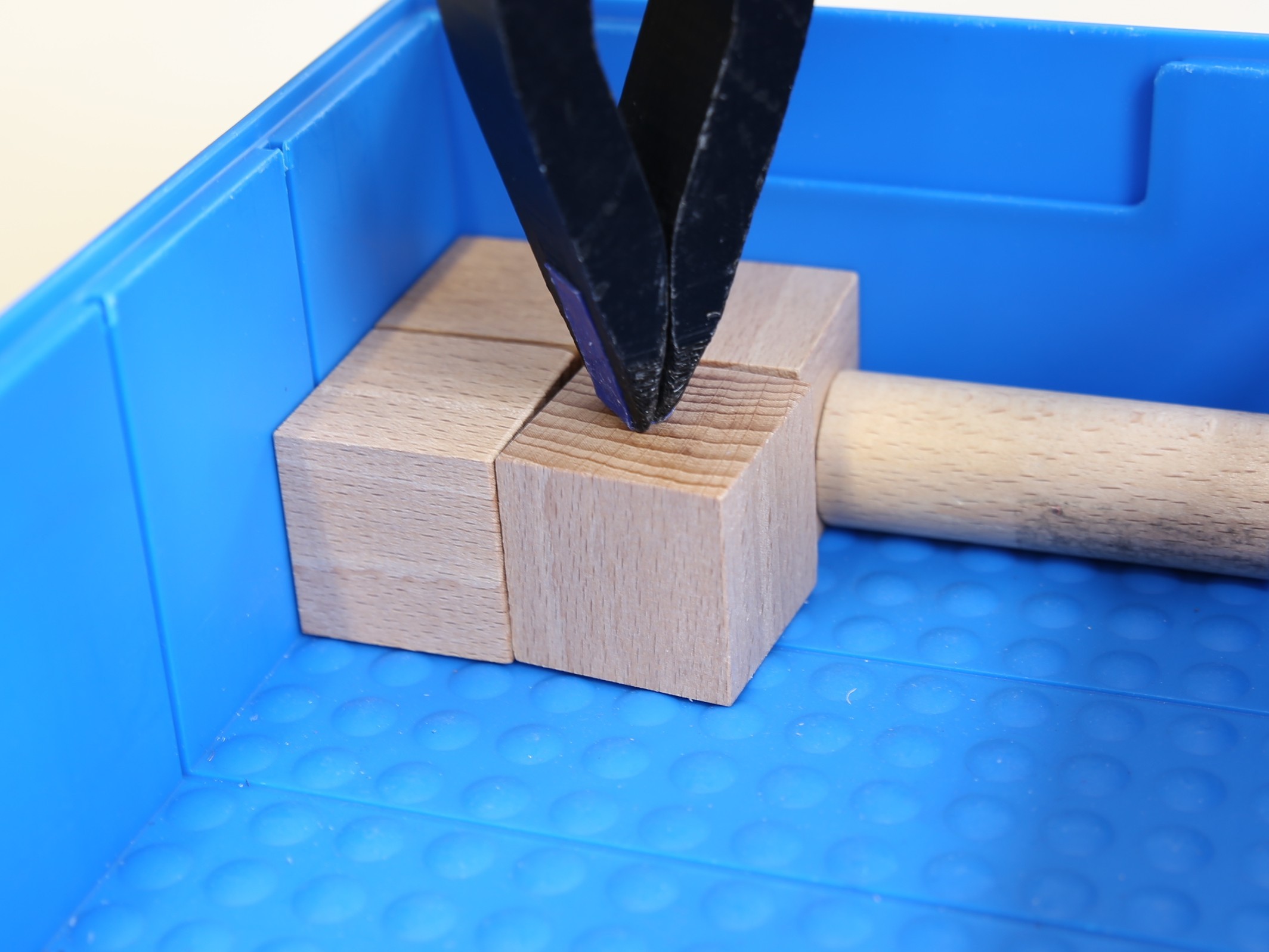}};
	\node[] (2) at (3.2, -0.7) {\includegraphics[trim=300 100 300 100, clip, width=0.29\linewidth]{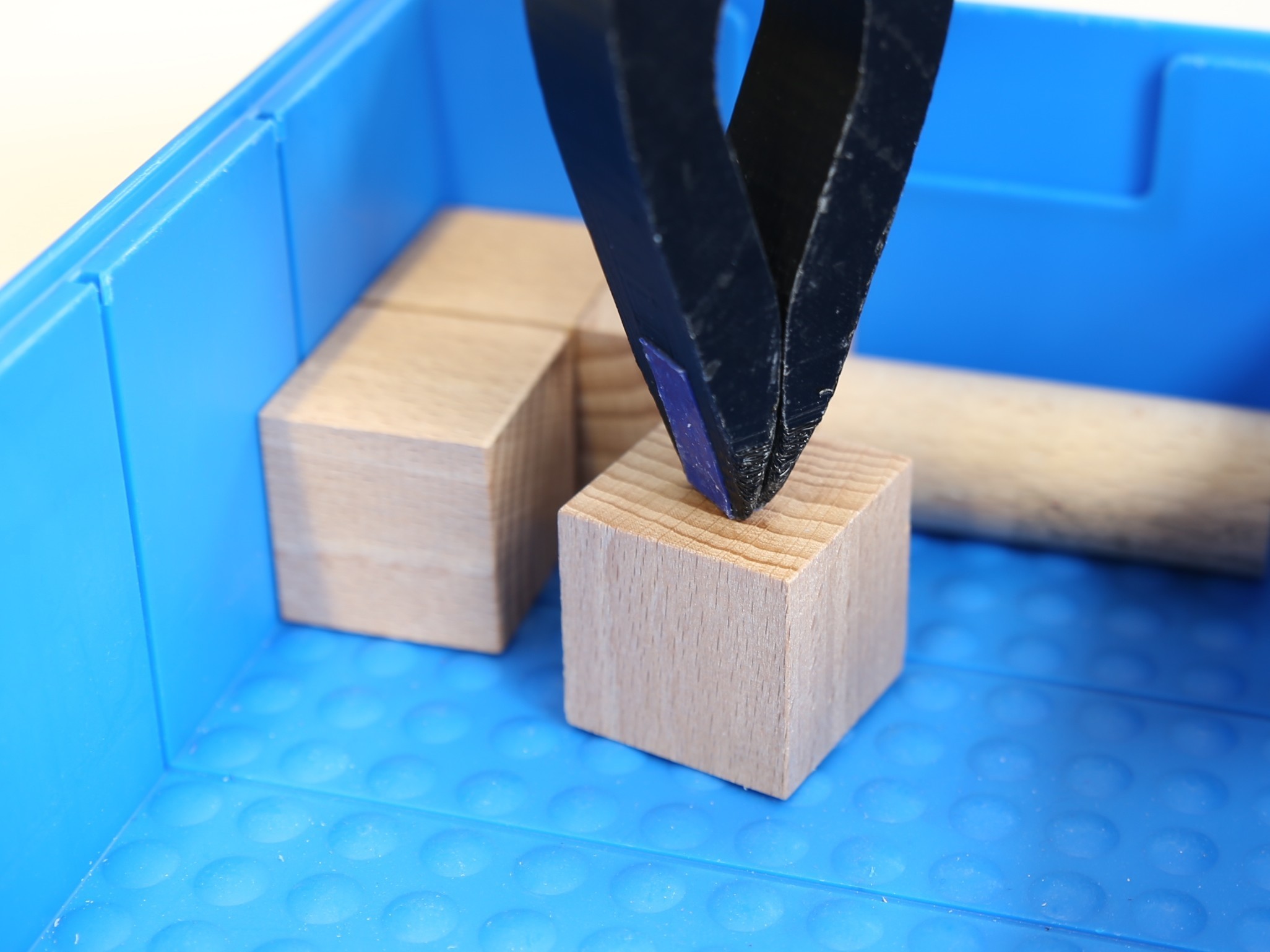}};
	\node[] (3) at (6.2, -0.7) {\includegraphics[trim=300 100 300 100, clip, width=0.29\linewidth]{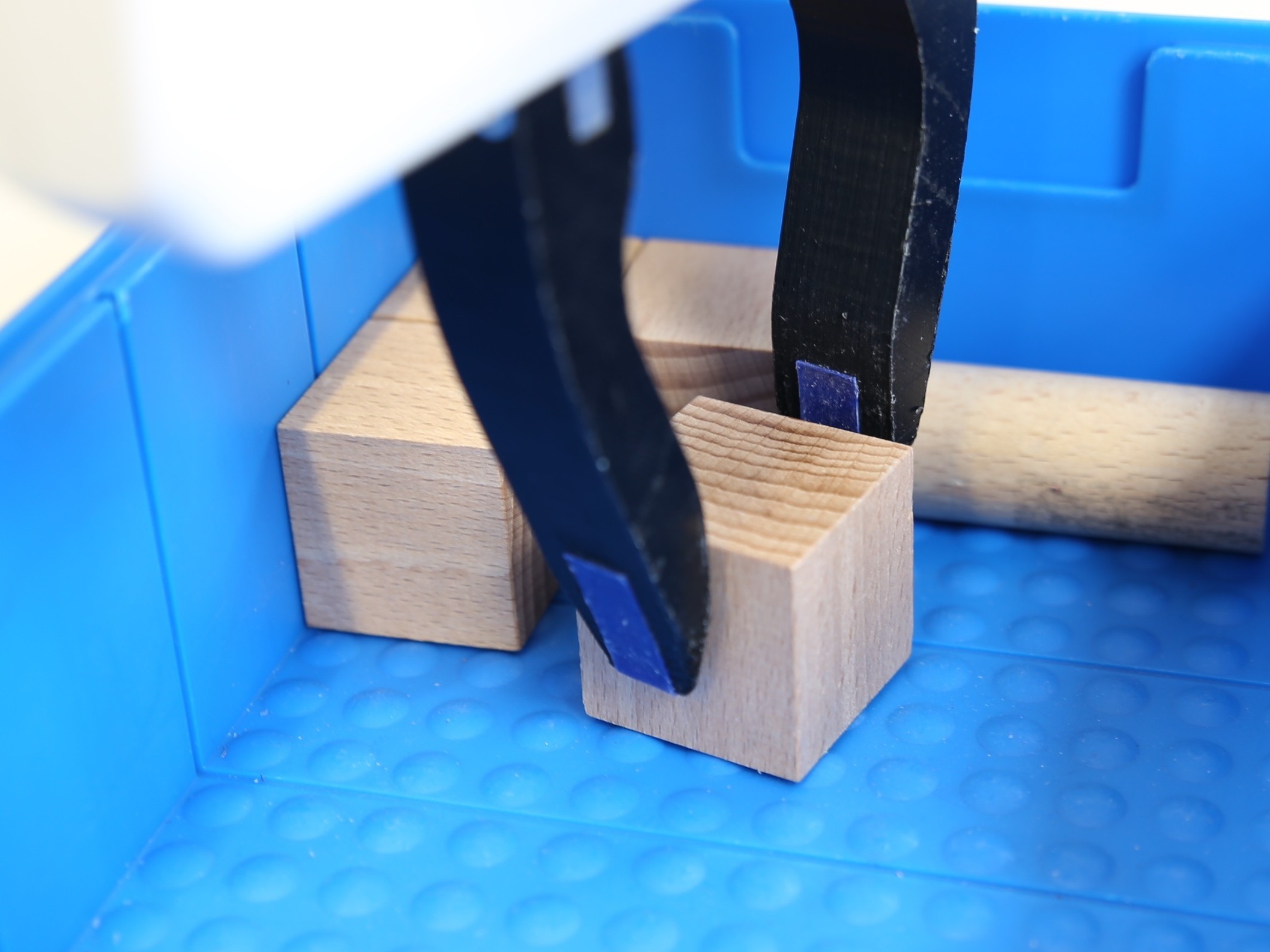}};
	
	\draw [thick, ->] (1.east) to (2.west);
	\draw [thick, ->] (2.east) to (3.west);

	\node[draw, rounded corners, fill=white, minimum height=18, text height=1.5ex, text depth=.25ex] at (1.9, -1.4) {1. Shift};
	\node[draw, rounded corners, fill=white, minimum height=18, text height=1.5ex, text depth=.25ex] at (5.1, -1.4) {2. Grasp};
\end{tikzpicture}

	\caption{Our setup of a Franka robotic arm including the standard force-feedback gripper (a), an Ensenso depth camera (b), custom 3D-printed gripper jaws with anti-slip tape (c), and two industrial bins with objects (d). The robot learns first grasping (2) and then shifting objects in order to explicitly increase grasp success (1).}
	\label{fig:overall-system}
\end{figure}
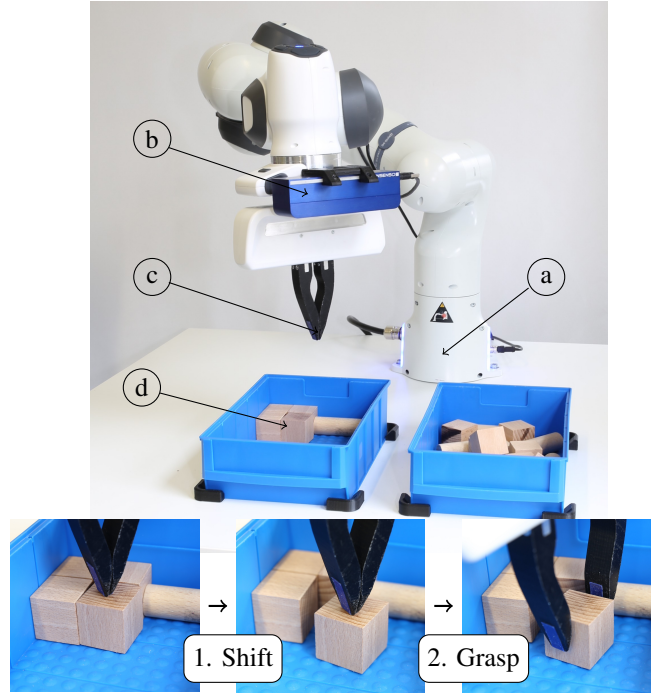

Our work is structured as follows: First, we present a vision-based algorithm for learning the most rewarding pose for applying object manipulation primitives. In our case, we define five primitives: Three for grasping at different gripper widths and two for shifting. Second, we use that approach to learn grasping by estimating the grasp probability at a given pose. Third, we derive both a grasping-dependent reward function as well as a training procedure for shifting. This way, sparse rewards of the grasp success can be bypassed for more data-efficient learning of shifting. Fourth, we present a robotic system (Fig.~\ref{fig:overall-system}) which learns the industrial task of bin picking. Beyond the capabilities of the first two steps, the system is able to empty bins completely and achieves arbitrary grasp rates at the expense of \acf{PPH}. Furthermore, we evaluate the system's ability of grasping novel objects from non-graspable positions.

\section{RELATED WORK}

Object manipulation and in particular grasping are well-researched fields within robotics. Bohg et al.~\cite{bohg_data-driven_2014} differentiate between analytical and data-driven approaches to grasping. Historically, grasp synthesis was based on analytical constructions of force-closure grasps \cite{ferrari_planning_1992}. In comparison, data-driven approaches are defined by sampling and ranking possible grasps. Popular ranking functions include classical mechanics and model-based grasp metrics \cite{miller_graspit!_2004, ferrari_planning_1992}. As modeling grasps itself is challenging, even more complex interactions like motion planning of pre-grasping actions were studied less frequently. Within this scope, Dogar et Srinivasa~\cite{dogar_push_2010} combined pushing and grasping into a single action, enabling them to grasp more cluttered objects from a table. Chang et al.~\cite{chang_planning_2010} presented a method for rotating objects to find more robust grasps for transport tasks.

In recent years, the progress of machine learning in computer vision enabled robot learning based on visual input \cite{finn_deep_2016}. As most approaches, in particular deep learning, are limited by its data consumption, data generation becomes a fundamental challenge. Possible solutions were supposed: First, training in simulation with subsequent sim-to-real transfer showed great results for grasp quality estimation \cite{mahler_dex-net_2016, bousmalis_using_2017}. However, as contact forces are difficult to simulate, training of more complex object interactions for pre-grasping manipulation might be challenging. Second, imitation learning deals with integrating expert knowledge by observing demonstrations \cite{pastor_online_2011}. Third, training of a robot using real-world object manipulation in a self-supervised manner showed great results for generalizing to novel objects \cite{pinto_supersizing_2016}. Levine et al.~\cite{levine_learning_2016} improved the grasp rate to \SI{82.5}{\%}, at the cost of upscaling the training to a multitude of robots for 2 months. Data consumption of learning for robotic grasping can be minimized by utilizing space invariances and improving the data exploration strategy \cite{berscheid_improving_2019}.

More recently, robot learning of manipulation skills are formulated as \ac{RL} problems. For differentiation, we find the action space either to be discrete, defined by a few motion primitives \cite{berscheid_improving_2019, zeng_learning_2018}, or continuous as a low-level control \cite{kalashnikov_scalable_2018, quillen_deep_2018}. While the latter allows for an end-to-end approach for general grasping strategies, it comes with the cost of very sparse rewards and a high data consumption. Kalashnikov et al.~\cite{kalashnikov_scalable_2018} trained an expensive multi-robot setup for \num{580000} grasp attempts, resulting in an impressive grasp rate of \SI{96}{\%} for unknown objects. Furthermore, the robots implicitly learned pre-grasping manipulation like singularization, pushing and reactive grasping. In contrast, Zeng et al.~\cite{zeng_learning_2018} introduced a pushing motion primitive and learned grasping and pushing in synergy by rewarding the sparse grasp success. Using significant less training data than \cite{kalashnikov_scalable_2018}, their robot was able to clear a table from tightly packed objects.

\section{SHIFT OBJECTS FOR GRASPING}

\Acf{RL} provides a powerful framework for robot learning. We introduce a \ac{MDP} $(\mathcal{S}, \mathcal{A}, T, r, p_0)$ with the state space $\mathcal{S}$, the action space $\mathcal{A}$, the transition distribution $T$, the reward function $r$ and the initial configuration $p_0$. 
Similar to other data-driven approaches, \ac{RL} is limited by its challenging data consumption, and even more so for time-dependent tasks including sparse rewards. For this reason, we reduce our process to a single time step. Then, a solution to this \ac{MDP} is a policy $\pi: \mathcal{S} \mapsto \mathcal{A}$ mapping the current state $s \in \mathcal{S}$ to an action $a \in \mathcal{A}$.

\subsection{Spatial Learning of Object Manipulation}

Given the visual state space $\mathcal{S}$, let $s$ denote the orthographic depth image of the scene. We simplify the action space to four parameters $(x, y, a, d) \in \mathcal{A} = \mathbb{R}^{3} \times \mathbb{N}$ in the planar subspace. The spatial coordinates $(x, y, a)$ are given in the image frame, using the usual $x$- and $y$-coordinate and the rotation $a$ around the axis $z$ orthogonal to the image frame. While the relative transformation between the camera frame and \ac{TCP} needs to be known, the absolute extrinsic calibration is learned. To get a full overview image of the object bin, we set the remaining angles $b=c=0$ resulting in planar object manipulation. The fourth parameter $d$ corresponds to the index within the discrete set of motion primitives $\mathcal{M}$.

\begin{figure}[t]
	\centering
	\vspace{1.6mm}
\begin{tikzpicture}[scale=0.69, inner sep=0pt, outer sep=2pt, line join=round, font=\scriptsize]
	\newcommand{\layer}[6][] {
		\draw[#1] (#2,-#3 / 2,-#4 / 2) -- ++(0,0,#4) -- ++(0,#3,0) -- ++(0,0,-#4) -- cycle;
		\draw[#1] (#2,-#3 / 2,-#4 / 2) -- +(#5,0,0);
		\draw[#1] (#2,#3 / 2,-#4 / 2) -- +(#5,0,0);
		\draw[#1] (#2,-#3 / 2,#4 / 2) -- +(#5,0,0) node [below=0.2cm, midway, align=center]{#6};
		\draw[#1] (#2,#3 / 2,#4 / 2) -- +(#5,0,0);
		\draw[#1] (#2 + #5,-#3 / 2,-#4 / 2) -- ++(0,0,#4) -- ++(0,#3,0) -- ++(0,0,-#4) -- cycle;
	}
	
	\newcommand{\window}[9][] {
		\draw[thin] (#2,-#3 / 2 + #6,-#4 / 2 + #7) -- ++(0,0,#4) -- ++(0,#3,0) -- ++(0,0,-#4) -- cycle;
		\draw[thin] (#2,-#3 / 2 + #6,-#4 / 2 + #7) -- ++(#5,0,0) -- ++(0,#3,0) -- ++(-#5, 0, 0) --cycle;
		\draw[thin] (#2,-#3 / 2 + #6,#4 / 2 + #7) -- ++(#5,0,0) -- ++(0,#3,0) -- ++(-#5, 0, 0) --cycle;
		\draw[thin] (#2 + #5,-#3 / 2 + #6,-#4 / 2 + #7) -- ++(0,0,#4) -- ++(0,#3,0) node [left=#1, midway]{#9} -- ++(0,0,-#4) node [above=#1, midway]{#8} -- cycle;
	}
	
	\newcommand{\calc}[7][] {
		\draw[#1, thin] (#2,-#3 / 2 + #6,-#4 / 2 + #7) -- ++(0,0,#4) -- ++(0,#3,0) -- ++(0,0,-#4) -- cycle;
		\draw[#1, thin, dashed] (#2,-#3 / 2 + #6,-#4 / 2 + #7) -- #5;
		\draw[#1, thin, dashed] (#2,#3 / 2 + #6,-#4 / 2 + #7) -- #5;
		\draw[#1, thin, dashed] (#2,-#3 / 2 + #6,#4 / 2 + #7) -- #5;
		\draw[#1, thin, dashed] (#2,#3 / 2 + #6,#4 / 2 + #7) -- #5;
	}
	
	\layer[thick]{0}{3}{3}{0}{1}
	\window[0.1cm]{0}{0.6}{0.6}{0.08}{0.5}{-0.3}{5}{5}
	\calc{0.08}{0.6}{0.6}{(1.4, 0.5, -0.3)}{0.5}{-0.3}
	
	\layer[thick]{1.4}{2.8}{2.8}{0.3}{32 \\ Stride $(2, 2)$, BN \\ Dropout $(0.4)$}
	\window[0.1cm]{1.4}{0.6}{0.6}{0.3}{0.5}{-0.3}{5}{5}
	\calc[]{1.4 + 0.3}{0.6}{0.6}{(3.1, -0.2, -0.1)}{0.5}{-0.3}
	
	\layer[thick]{3.1}{1.3}{1.3}{0.4}{48 \\ BN \\ Dropout $(0.4)$}
	\window[0.1cm]{3.1}{0.6}{0.6}{0.4}{-0.2}{-0.1}{5}{5}
	\calc[]{3.1 + 0.4}{0.6}{0.6}{(4.7, -0.1, 0)}{-0.2}{-0.1}
	
	\layer[thick]{4.7}{1.1}{1.1}{0.6}{64 \\ BN \\ \qquad\qquad Dropout $(0.3)$}
	\window[0.3cm]{4.7}{0.7}{0.7}{0.6}{-0.1}{0}{6}{6}
	\calc[]{4.7 + 0.6}{0.7}{0.7}{(6.2, 0, 0)}{-0.1}{0}
	
	\layer[thick]{6.4}{0.4}{0.4}{1.3}{142 \\ BN \\ Dropout $(0.3)$}
	\window[0.3cm]{6.4}{0.1}{0.1}{1.3}{0}{0}{1}{1}
	\calc[]{6.4 + 1.3}{0.1}{0.1}{(8.7, 0, 0)}{0}{0}
	
	\layer[thick]{8.7}{0.4}{0.4}{1.0}{128 \\ Dropout $(0.3)$}
	\window[0.3cm]{8.7}{0.1}{0.1}{1.0}{0}{0}{1}{1}
	\calc[]{8.7 + 1.0}{0.1}{0.1}{(10.8, 0, 0)}{0}{0}
	
	\layer[thick]{10.7}{0.4}{0.4}{0.1}{$\vert \mathcal{M} \vert$}
\end{tikzpicture}
	
	\caption{Our fully-convolutional \acf{NN} architecture, making use of both \acf{BN} and dropout for a given motion primitive set $\mathcal{M}$.}
	\label{fig:nn-architecture}
\end{figure}
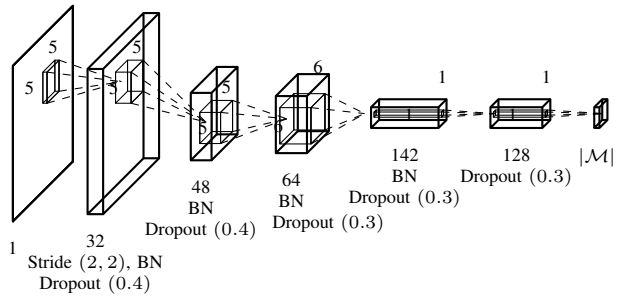

\begin{figure*}[t]
	\centering
	\vspace{1.6mm}
	
	\newcommand{\subfiguresize}{0.30\linewidth}
	\newcommand{\graphicssize}{0.44\linewidth}
	
	\begin{subfigure}[t]{\subfiguresize}
		\centering
		\includegraphics[width=\graphicssize]{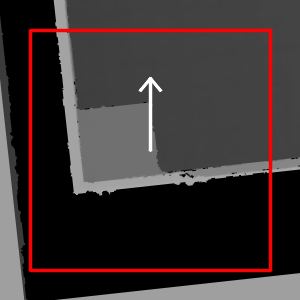}
		~
		\includegraphics[width=\graphicssize]{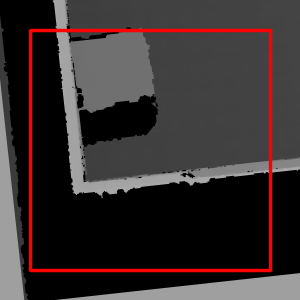}

		\caption{$\hat{\psi}_{w} = \SI{0.046}{\%}$, $\hat{\psi}_{w}^\prime = \SI{93.2}{\%}$}
	\end{subfigure}
	\quad
	\begin{subfigure}[t]{\subfiguresize}
		\centering
		\includegraphics[width=\graphicssize]{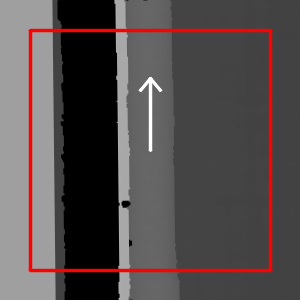}
		~
		\includegraphics[width=\graphicssize]{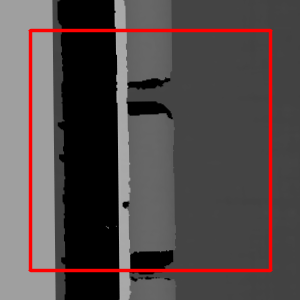}

		\caption{$\hat{\psi}_{w} = \SI{14.7}{\%}$, $\hat{\psi}_{w}^\prime = \SI{75.9}{\%}$}
	\end{subfigure}
	\quad
	\begin{subfigure}[t]{\subfiguresize}
		\centering
		\includegraphics[width=\graphicssize]{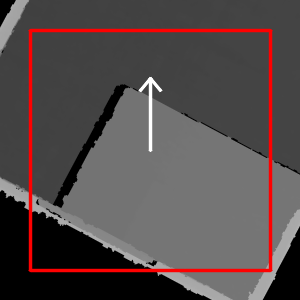}
		~
		\includegraphics[width=\graphicssize]{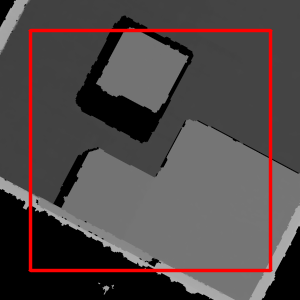}

		\caption{$\hat{\psi}_{w} = \SI{40.2}{\%}$, $\hat{\psi}_{w}^\prime = \SI{97.7}{\%}$}
	\end{subfigure}
	
	\vspace{2mm}
	
	\begin{subfigure}[t]{\subfiguresize}
		\centering
		\includegraphics[width=\graphicssize]{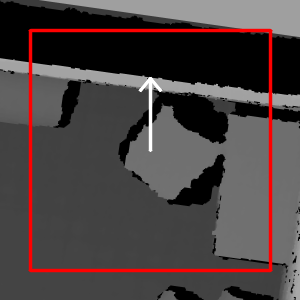}
		~
		\includegraphics[width=\graphicssize]{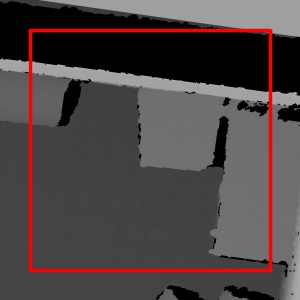}

		\caption{$\hat{\psi}_{w} = \SI{92.1}{\%}$, $\hat{\psi}_{w}^\prime = \SI{43.9}{\%}$}
	\end{subfigure}
	\quad
	\begin{subfigure}[t]{\subfiguresize}
		\centering
		\includegraphics[width=\graphicssize]{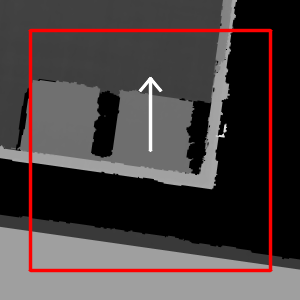}
		~
		\includegraphics[width=\graphicssize]{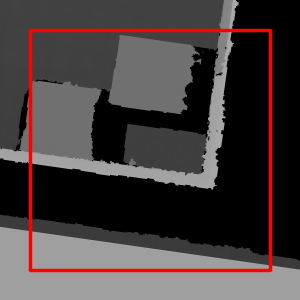}

		\caption{$\hat{\psi}_{w} = \SI{90.5}{\%}$, $\hat{\psi}_{w}^\prime = \SI{95.8}{\%}$}
	\end{subfigure}
	\quad
	\begin{subfigure}[t]{\subfiguresize}
		\centering
		\includegraphics[width=\graphicssize]{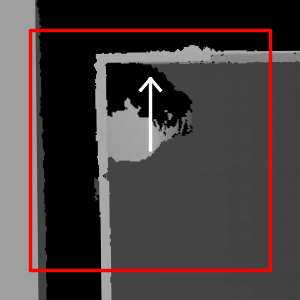}
		~
		\includegraphics[width=\graphicssize]{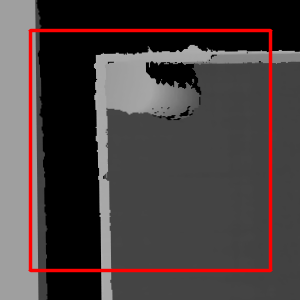}

		\caption{$\hat{\psi}_{w} = \SI{90.8}{\%}$, $\hat{\psi}_{w}^\prime = \SI{45.8}{\%}$}
	\end{subfigure}
	
	\caption{Examples of depth images before (left) and after (right) an applied motion primitive. The maximal grasp probability within the red window $\hat{\psi}_{w}$ before and $\hat{\psi}_{w}^\prime$ after are given below; their difference is then estimated by a fully-convolutional neural network.}
	\label{fig:example-image-difference}
\end{figure*}

The policy $\pi(s) = \sigma \circ Q(s, a)$ is split into an action-value function $Q$ and a selection function $\sigma$. $Q(s, a)$ estimates the reward $r$ for an action $a$ given an orthographic depth image $s$. We introduce a sliding window $s^\prime \subset s$ that crops the orthographic image at the given translation $(x, y)$ and rotation $(a)$. The $x$-$y$-translation is implemented efficiently as a fully convolutional \ac{NN}, the rotation parameters by applying the \ac{NN} on multiple pre-rotated images. The motion primitive set $\mathcal{M}$ is calculated by the number of output channels of the last convolutional layer. Fig.~\ref{fig:nn-architecture} shows the detailed architecture of the used \ac{NN}. The training output has a size of $(1 \times 1 \times \vert \mathcal{M} \vert)$, corresponding to the size of the motion primitive set $\mathcal{M}$. During inference, the \ac{NN} calculates an output of size $(40 \times 40 \times \vert \mathcal{M} \vert)$, corresponding to the $(x, y, d)$ parameters. For the rotation $a$, the input image is pre-transformed and the \ac{NN} is recalculated for $20$ angles. This way, \num{32000} reward estimations are calculated for each motion primitive. Overall, the \ac{NN} approximates the action-value function $Q$ for a discrete set of actions $a$ within four dimensions.

The selection function $\sigma$ maps the four dimensional tensor of reward predictions $r$ to an action $a$. In \ac{RL}, the greedy strategy using $\argmax_a Q(s, a)$ is commonly used for sufficiently trained systems. By returning one of the $N$ maximum elements uniformly, we integrate a stochastic component so that possible manipulation failures are not repeated. The indices of the selected element at $(x, y, a)$ are then transformed into the camera frame using the intrinsic calibration. The height $z$ is read trivially from the depth image at position $(x, y)$, adding an fixed height offset for each specific motion primitive $d$. Finally, the algorithm outputs the pose $\left(x, y, z, a, b=\text{const}, c=\text{const} \right)$ of a defined motion primitive $d$ with an estimated reward $Q$.

\subsection{Action Space Exploration}
\label{subsec:space-exploration}
\newcommand{\minus}{\scalebox{0.75}[1.0]{$-$}}

As the design of an exploration strategy is a key to fast and data-efficient learning for robotic grasping \cite{berscheid_improving_2019}, we introduce high-level strategies generalized for further object manipulation. We strictly divide between exploration (training) and exploitation (application) phase. Without prior information, the system explores the environment by sampling continuously and uniformly random poses within the hull of the action space $\mathcal{A}$. Let $\varepsilon$ define the fraction of random samples. Due to our off-policy algorithm, we are able to combine an $\varepsilon$-based strategy with the following set of high-level strategies:
\begin{enumerate}
\item \textbf{Maximize self-information} corresponding to $\max_x \minus\log \tilde{P}(x)$ with the estimated probability mass function $\tilde{P}(x)$. For manipulation tasks, actions with high absolute rewards are usually more seldom. Therefore, the action with the maximum reward estimation $\max_a \vert Q(s, a) \vert$ should be chosen. This conforms with the common $\varepsilon$-greedy strategy. In comparison, we find that sampling corresponding to $p(a)\sim\vert Q(s, a) \vert$ yields a more extensive exploration.
\item \textbf{Minimize uncertainty of prediction} given by $\max_a \text{Var}\left[ Q(s, a) \right]$. In \ac{RL}, this is usually added to the action-value function itself (for exploitation), leading to the common \Ac{UCB} algorithm. We approximate the Bayesian uncertainty of our \ac{NN} using Monte-Carlo dropout for variance sampling \cite{gal_dropout_2015}.
\item \textbf{Minimize uncertainty of outcome} for binary rewards $r = \lbrace 0, 1 \rbrace$ by choosing $\min_a \vert Q(s, a) - \frac{1}{2} \vert$. The system is not able to predict the outcome for those actions reliably, e.g.\ due to missing information or stochastic physics.
\end{enumerate}

\subsection{Learning for Grasping}

A major contribution of our work is making one skill (shifting) explicitly dependent on the other (grasping). This way, we bypass the problem of sparse-rewards in time-dependent \acp{MDP}. Besides faster and more data-efficient training, this also allows to learn the skills successively.

Therefore, we can reuse successful approaches of learning for grasping \cite{berscheid_improving_2019} and focus on pre-grasping manipulation. Briefly, we define the set of motion primitives $\mathcal{M}$ as gripper clamping actions starting from three different pre-shaped gripper widths. The robot's trajectory is given by the grasp pose $(x, y, z, a, b, c)$ and its approach vector parallel to the gripper jaws. If the robot detects a collision with its internal force-sensor, the robot retracts a few millimeters and closes the gripper. Then, the object is lifted and the robot moves to a random pose above the filing bin. Then, the grasp success is measured using the force-feedback of the gripper. We define the binary reward function
\begin{align}
r_g(s) &= \begin{cases} 
	1 & \text{if grasp and filing successful,} \\
	0 & \text{else.}
\end{cases}
\end{align}
For binary rewards, the grasping action-value function $Q_g(s, a)$ can be interpreted as a grasp probability $\psi$. We train a \ac{NN} mapping the image $s$ to $\psi$ and use it for: (1) estimating the grasp probability at a given position $(x, y, a)$, (2) calculating the best grasp $(x, y, a, d)$, (3) calculating the maximum grasp probability in the entire bin $\hat{\psi}$, and (4) calculating the maximum grasp probability $\hat{\psi}_{w}(s, x, y, a)$ in a window with a given side length centered around a given pose $(x, y, a)$.

\subsection{Learning for Shifting}

We integrate prior knowledge about the relationship between shifting and grasping by making the reward function for shifting explicitly dependent on the grasping probability. More precisely, the system predicts the influence of a motion primitive on the maximum grasp probability $\hat{\psi}$. We train a second \ac{NN} using the reward function
\begin{align}
r_s(s) &= \frac{1}{2} \left( \hat{\psi}_{w}(s^{\prime}, x, y, a) - \hat{\psi}_{w}(s, x, y, a) + 1 \right)
\end{align}
mapping the image $s$ to the difference of the maximum grasping probability in a window $\hat{\psi}_{w}(s, x, y, a)$ before $s$ and after $s^{\prime}$ the manipulation primitive. Therefore, depth images before and after the shifting attempt are recorded and applied to the grasp probability \ac{NN}. Additionally, the reward is re-normalized to $r_s \in \left[ 0, 1 \right]$. The window $w$ is approximately \SI{50}{\%} larger than for the grasping action, the latter corresponding roughly to the maximum gripper width. In contrast to the grasping reward function $r_g \in \lbrace 0, 1 \rbrace$, estimating shift rewards is a regression task. We further denote the estimated reward for shifting $Q_s(s, a)$ as $\rho$. The \ac{NN} is trained optimizing the mean squared loss between the predicted and actual reward $\rho$ and $r_s$. 

We define two motion primitives for pre-grasping object manipulation. In both cases, the gripper closes completely and approaches the shift pose parallel to its gripper jaws. If a collision occurs, the approach motion stops. Then, the robot moves either \SI{30}{mm} in positive x-direction or positive y-direction of the gripper frame. Those two motion primitives are distinct as the gripper is asymmetric.

Since data generation is the limiting factor in deep \ac{RL}, it is important that training is self-supervised and requires as little human intervention as possible. To guarantee a continuous training, the system first estimates the overall maximum grasping probability $\hat{\psi}$. If $\hat{\psi} < 0.2$, the system tries to increase the grasping probability until $\hat{\psi} \geq 0.8$ is reached. Then, the system tries to decrease the maximum grasping probability until $\hat{\psi} < 0.2$ again. This is done by exploring the negative action-value-function $\minus Q_s(s, a)$ while keeping the selection function $\sigma$ constant. Training started with a single object in the bin, further ones were added over time.

\subsection{Combined Learning and Inference}

\begin{figure}
	\centering
	\vspace{1.6mm}
\begin{tikzpicture}
  \node[draw, rounded corners] (grasp-inf) at (0.0, -0.6) {Estimate max grasp probability $\hat{\psi}$};
  \node[draw, rounded corners] (shift-inf) at (-1.3, -2.0) {Estimate max shift reward $\hat{\rho}$};
  \node[draw, rounded corners] (grasp) at (2.1, -2.0) {Grasp};
  
  \node[draw, rounded corners] (shift) at (-2.6, -3.4) {Shift};
  \node[draw, rounded corners] (empty) at (0.1, -3.4) {Bin empty};
  
  \node[] () at (-1.9, -1.3) {$\hat{\psi} < \psi_g$};
  \node[] () at (2.7, -1.3) {$\hat{\psi} \geq \psi_g$};

  \node[] () at (-3.0, -2.7) {$\hat{\rho} \geq \rho_s$};
  \node[] () at (0.5, -2.7) {$\hat{\rho} < \rho_s$};

  \draw[->] (grasp-inf) to (grasp.north);
  \draw[->] (grasp-inf) to (shift-inf.north);
  \draw[->] (shift-inf) to (shift.north);
  \draw[->] (shift-inf) to (empty.north);
  \draw[->] (shift) |- (-4.0, -3.9) |- (-4.0, 0.05) -| (grasp-inf.north);
  \draw[->] (grasp) |- (4.0, -2.51) |- (4.0, 0.05) -| (grasp-inf.north);
\end{tikzpicture}
	\caption{State diagram of the combined grasping and shifting controller; common threshold parameters are $\psi_g \approx \num{0.75}$ and $\rho_s \approx \num{0.6}$.}
	\label{fig:inference-state}
\end{figure}
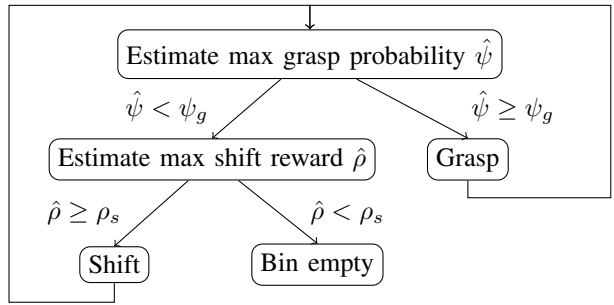

For the task of bin picking, grasping and shifting needs to be combined into a single controller. Beside inference itself, combined learning also enables matching data distributions for training and application. Firstly, let $\psi_g$ be a threshold probability deciding between a grasping and a shifting attempt. Secondly, let $\rho_s$ denote a threshold between a shift attempt and the assumption of an empty bin. As shown in Fig.~\ref{fig:inference-state}, the system first infers the maximum grasping probability $\hat{\psi}$. If $\hat{\psi}$ is higher than $\psi_g$ the robot grasps, else it evaluates the shifting \ac{NN} and estimates the maximum shifting reward $\hat{\rho}$. If $\hat{\rho}$ is larger than $\rho_s$ the robot shifts and restart the grasp attempt, else it assumes the bin to be empty.
$\psi_g$ can be interpreted as a high-level parameter corresponding to the system's cautiousness for grasp attempts as well as its readiness to rather increase the grasp probability by shifting.

\section{EXPERIMENTAL RESULTS}

\begin{table*}[ht]
	\centering
	\vspace{1.4mm}
	\caption{\Acf{PPH}, grasp rate, and shifts per grasp in different bin picking scenarios. In the experiment, the robot grasped $n$ objects out of a bin with $m$ objects without replacement. The random grasp rate was $\approx$ \SI{3}{\%}.}
\begin{tabular}{|l|c|c|c|c|}
	\hline
	\textbf{$n$ out of $m$} & \textbf{\Acf{PPH}} & \textbf{Grasp Rate} & \textbf{Shifts per Grasp} & \textbf{Grasp Attempts} \\
	\hline
	1 out of 1 & \num{323 \pm 3} & \SI{100}{\%} & \si{0.02 \pm 0.01} & \num{100} \\
	1 out of 1 (non-graspable position) & \num{170 \pm 3} & \SI{98.2 \pm 1.8}{\%} & \num{1.02 \pm 0.02} & \num{56} \\
	\hline
	10 out of 10 & \num{272 \pm 6} & \SI{98.4 \pm 1.1}{\%} & \num{0.10 \pm 0.02} & \num{122} \\
	10 out of 20 & \num{299.6 \pm 1.4} & \SI{100}{\%} & \num{0} & \num{120} \\
	20 out of 20 & \num{274 \pm 3} & \SI{98.4 \pm 1.0}{\%} & \num{0.07 \pm 0.01} & \num{122} \\
	\hline
\end{tabular}
	\label{tab:results}
\end{table*}

Our experiments are performed on the Franka Panda robotic arm with the standard gripper as seen in Fig.~\ref{fig:overall-system}. The Ensenso N10 stereo camera is mounted on the robot's flange. For all state observations, the camera is positioned above the bin looking down vertically, keeping transformations between robot, camera and world coordinates fixed. We designed and printed custom gripper jaws and attached common household anti-slip silicone roll to the robot's fingertips. Otherwise, the friction between the fingers and most objects would not be sufficient for shifting. The robot uses an Intel Core i7-8700K processor and two NVIDIA GeForce GTX 1070 Ti for computing. Inferring the \ac{NN} takes around \SI{10}{ms} on a single GPU\@, calculating the next action including image capturing takes less than \SI{100}{ms}. The source code, trained models and a supplementary video showing our experimental results are published at \url{https://github.com/pantor/learning-shifting-for-grasping}.

\subsection{Data Recording}

In bin picking, shifting objects becomes either necessary because of static obstacles like the bin or dynamic obstacles like other objects. To enforce the first case, we use a small bin with the size of $\SI{18}{cm} \times \SI{28}{cm}$. The latter case is emphasized by using cubes as the most suitable shape for blocking grasps among themselves. Additionally, we train with wooden cylinders as second object primitives. For our final data set, we trained \num{25000} grasps in around \SI{100}{h}. For learning to shift, we recorded \num{2500} attempts in around \SI{9}{h}. We find that grasping and shifting need different amounts of training time; separate training allows for easy stopping at an appropriate success measure and an overall data-efficient recording. Regarding exploration strategies, we maximized \textit{self-information} around \SI{60}{\%}, minimized the \textit{uncertainty of prediction} around \SI{20}{\%}, and minimized the \textit{uncertainty of outcome} around \SI{5}{\%} with decreasing $\varepsilon$, as well as using random actions during the first \SI{15}{\%} of recording. Furthermore, we generated data using the combined training approach for the last \SI{10}{\%} of the training time. For robust and low-noise calculations of $\rho$, the grasping probability $\psi$ needs to be trained reliably for at least \num{15000} grasp attempts. 

\subsection{Shifting Objects}

\begin{figure}[!b]
	\centering
	
	\begin{subfigure}[t]{0.82\linewidth}
		\centering
		\includegraphics[trim=85 40 45 40, clip, width=\linewidth]{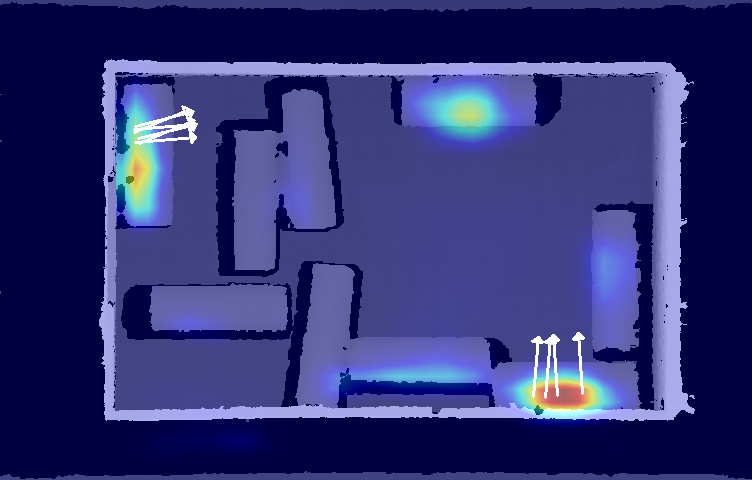}
	\end{subfigure}
	
	\vspace{1mm}
	
	\begin{subfigure}[t]{0.82\linewidth}
		\centering
		\includegraphics[trim=85 40 45 40, clip, width=\linewidth]{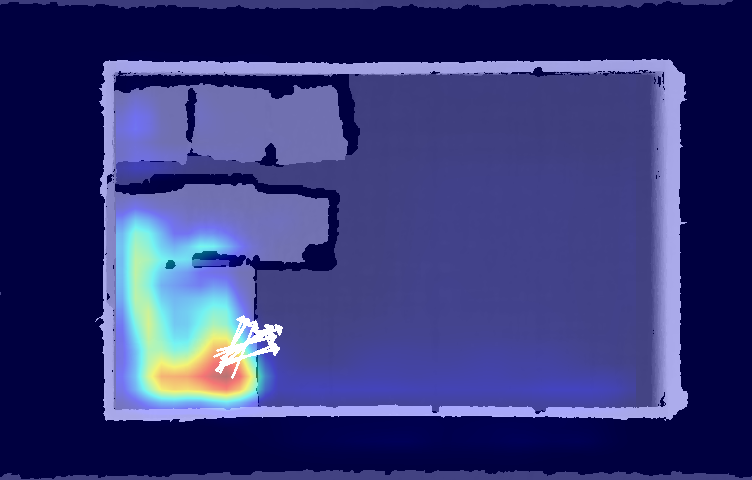}
	\end{subfigure}
	
	\caption{Examples of heat maps for shifting. The \ac{NN} predicts the maximum positive reward $\rho$ at a given pose $(x, y)$ for all rotations $a$ and motion primitives $d$. The rewards are shown from low (blue, $\rho = 0.5$) to high (red, $\rho = 1$), the direction is shown (white) for the ten highest rewards.}
	\label{fig:heatmap-shifting}
\end{figure}

The recorded shifting data set has a mean reward of $\num{0.521 \pm 0.112}$. The trained \ac{NN} achieves a cross validated mean squared loss of $\num{0.053}$, corresponding to a mean error of the grasp probability difference $\vert \hat{\psi}_w^\prime - \hat{\psi}_w \vert$ of around \SI{14}{\%}. Setting $\rho_s=0.6$ is a robust threshold for perceiving empty bins. The output of the \ac{NN} can be interpreted as a heat map over the original input image. Fig.~\ref{fig:heatmap-shifting} shows qualitatively that the system learned to shift objects either apart or away from the bin's edge for improved grasping. We denote non-graspable positions as object poses, where at least one shift is required before grasping. As given in Table~\ref{tab:results}, on average \SI{2}{\%} of shifts are not sufficient for grasping single objects from those positions.

\subsection{Picks Per Hour}

Shifting objects enabled the robot to empty the bin completely throughout the evaluation. For realistic scenarios like grasping $20$ objects out of a bin with $20$ objects without replacement, our robotic setup achieved \num{274 \pm 3}~\ac{PPH}. The fewer shifts per grasp are necessary for the bin picking scenario, the higher the \ac{PPH}. Using the grasp threshold parameter $\psi_g$, we can adapt the grasp strategy to be more cautious. Let the grasp rate be the number of grasp success over the total number of grasp attempts.
\begin{figure}[ht]
	\centering
	\vspace{1.6mm}

\begin{tikzpicture}[scale=0.86]
\pgfplotsset{
    xmin=0.15, xmax=1.0
}

\draw[dashed, black!60] (0, 5.42) -- (6.9, 5.42);
\draw[dashed, black!60] (4.8, 0) -- (4.8, 5.7);

\begin{axis}[
	axis y line*=left,
	ymin=0, ymax=1.05,
	xlabel=Grasp Threshold Parameter $\psi_g$,
	ylabel=Grasp Rate,
	legend pos=south west,
]
	\addplot[mark=x, red] table [y=grasprate, x=threshold]{figures/grasp-rate-threshold.txt};
	\label{plot_one}
\end{axis}

\begin{axis}[
	axis y line*=right,
	axis x line=none,
	ymin=120, ymax=320,
	ylabel=PPH,
	legend pos=south west,
]
	\addlegendimage{/pgfplots/refstyle=plot_one}\addlegendentry{Grasp Rate}
	\addplot[mark=*, blue] table [y=pph, x=threshold]{figures/grasp-rate-threshold.txt};
	\addlegendentry{PPH}
\end{axis}
\end{tikzpicture}

	\caption{Grasp rate and \acf{PPH} depending on the minimum grasp probability $\psi_g$ deciding between a grasp or shift. While the robot grasped 10 objects out of a bin with 10 objects without replacement, an optimal threshold $\psi_g \approx 0.75$ regarding \ac{PPH} was measured.}
	\label{fig:pph-grasp-threshold}
\end{figure}
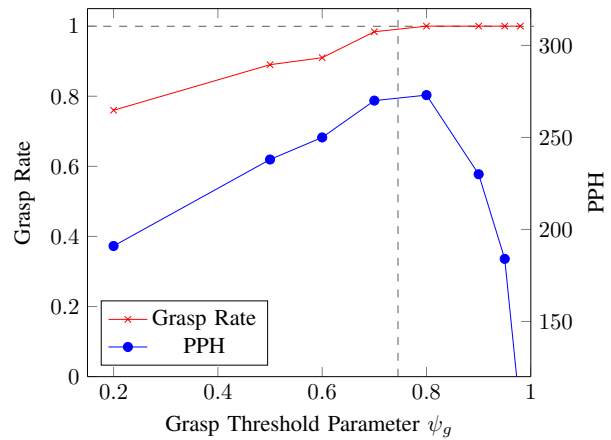
As expected, Fig.~\ref{fig:pph-grasp-threshold} shows that a higher $\psi_g$ results in an improved grasp rate. In particular, the system is able to achieve a \SI{100}{\%} grasp rate already for $\psi_g > 0.8$. For this reason, the grasp rate looses its significance as the major evaluation metric. Instead, we can directly optimize the industrially important metric of \acf{PPH}. At low $\psi_g$, frequent failed grasps take a lot of time. For high $\psi_g$, the improved grasp rate comes with the cost of more shifts. Since some shifts might be superfluous and worsen the \ac{PPH}, an optimal threshold has to exist. Fig.~\ref{fig:pph-grasp-threshold} confirms our expectation, resulting in an optimal threshold $\psi_g \approx 0.75$. Interestingly, the corresponding grasp rate is less than $1$.

\subsection{Generalization}

\begin{figure}
    \centering
    \vspace{1.6mm}
    \includegraphics[width=0.8\linewidth]{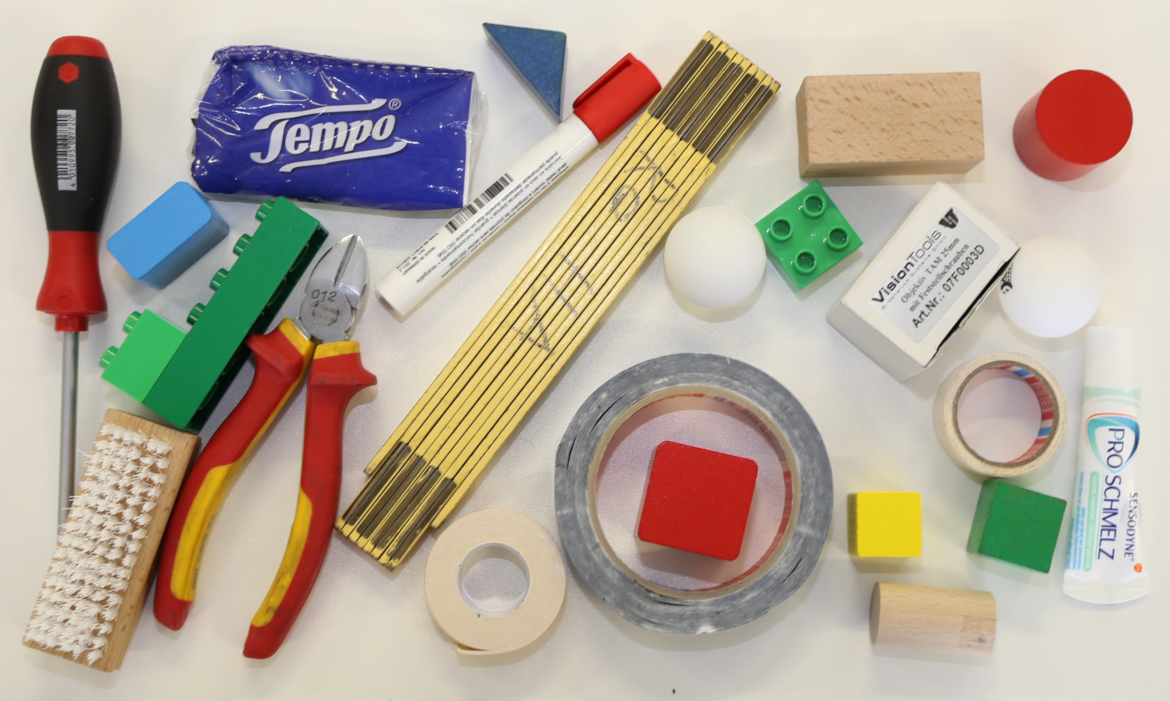}
    \caption{The object set for testing the system's ability to generalize to unknown objects. All objects can be placed within a bin so that they can not be grasped directly.}
    \label{fig:object-test-set}
\end{figure}

The ability to generalize to novel (unknown) objects is important for extending the range of applications. While training only with cylinders and cubes, we further evaluate the system on the object test set shown in Fig.~\ref{fig:object-test-set}. On average, the robot was able to grasp novel objects from non-graspable positions after \num{1.2 \pm 0.3} shifts (Table~\ref{tab:generalization}). Then, the robot achieved an average grasp rate of \SI{92.1 \pm 7.8}{\%}. As expected, we find a qualitative relation between the similarity to the training objects and the success rate of each test object. However, the most common cause of failure is missing depth information from the stereo camera and the resulting confusion by large black regions (e.g.\ shown in Fig.~\ref{fig:example-image-difference}). Furthermore, the shifting motion primitives are not equally suitable for each object. For example, the marker was usually shifted twice, as it did not roll far enough else. The brush did sometimes rebound back to its original position, which could be prevented by a larger shifting distance.

\begin{table}[ht]
	\centering
	\caption{Grasp rate and shifts per grasp for novel (unknown) objects from non-graspable positions.}
	\begin{tabular}{|l|c|c|c|}
	\hline
	\textbf{Object} & \textbf{Grasp Rate} & \textbf{Shifts} & \textbf{Grasp Attempts} \\
	\hline
	Brush & \SI{77}{\%} & \num{1.3} & 10 \\
	Cardboard box & \SI{94}{\%} & \num{1} & 15 \\
	Duplo bricks & \SI{91}{\%} & \num{1.3} & 10 \\
	Folding rule & \SI{83}{\%} & \num{1.1} & 10 \\
	Marker & \SI{100}{\%} & \num{1.9} & 10 \\
	Pliers & \SI{83}{\%} & \num{1.2} & 10 \\
	Screw driver & \SI{83}{\%} & \num{1.2} & 10 \\
	Shape primitives & \SI{98}{\%} & \num{1} & 25 \\
	Table tennis balls & \SI{100}{\%} & \num{1.1} & 10 \\
	Tape & \SI{92}{\%} & \num{1.5} & 10 \\
	Tissue wrap & \SI{100}{\%} & \num{1} & 10 \\
	Toothpaste & \SI{100}{\%} & \num{1.1} & 10 \\
	\hline
	& \SI{92 \pm 7}{\%} & \num{1.2 \pm 0.3} & \num{140} \\
	\hline
	\end{tabular}
	\label{tab:generalization}
\end{table}

\section{DISCUSSION AND OUTLOOK}

We presented a real-world solution for self-supervised learning of grasping and manipulating to improve expected grasp success. We find two implications of our work particularly interesting: First, emptying a bin completely is important for industrial applications. Second, we integrated prior knowledge into the learning algorithm by making the reward of one skill (shifting) dependent on the other (grasping). This way, we were able to bypass sparse rewards for data-efficient learning.

In contrast, both Kalashnikov et al.~\cite{kalashnikov_scalable_2018} and Zeng et al.~\cite{zeng_learning_2018} rewarded grasp success in a time-dependent manner. Additionally, we focused on bin picking scenarios of densely filled, industrial storage bins. For this task, we find multiple gripper openings, neglected by both \cite{kalashnikov_scalable_2018, zeng_learning_2018}, inevitable. While our approach is more similar to \cite{zeng_learning_2018}, we highlight four concrete improvements: First, we integrated multiple motion primitives into a single \ac{NN}. Second, our controller is able to change its readiness to assume risk. This way, our robot achieves arbitrary grasp rates, so that we can directly optimize in relation to \acf{PPH}. Third, while we find their contribution of training grasping and pushing in synergy necessary, it is not ideal on its own. By splitting both the rewards and training procedures for grasping and shifting, we incorporate different training complexities for different skills. Fourth, our algorithm is around an order of magnitude faster, resulting in increased \ac{PPH}. Regarding \cite{kalashnikov_scalable_2018}, the \ac{PPH} are fundamentally restricted by their repeating inference and motion steps.

As all data-driven methods benefit from a similar training and test data distribution, generalization can always be improved by training on more diverse object and scenario sets. While we showed that our approach is able to generalize to novel objects, possible limits of this ability should be further investigated. Moreover, our algorithm requires orthographic images for \ac{NN} input and therefore depth information. However, as viewing shadows or reflective surfaces limit depth availability for stereo cameras, we find the robustness against missing depth information to be a key part for further improving the robotic bin picking system.

\bibliographystyle{IEEEtran}
\bibliography{root}

\end{document}